\documentclass[letterpaper]{article} 
\usepackage{aaai25}  
\usepackage{times}  
\usepackage{helvet}  
\usepackage{courier}  
\usepackage[hyphens]{url}  
\usepackage{graphicx} 
\usepackage{natbib}  
\usepackage{caption} 
\usepackage{algorithm}
\usepackage{algorithmic}

\usepackage{newfloat}
\usepackage{listings}
\DeclareCaptionStyle{ruled}{labelfont=normalfont,labelsep=colon,strut=off} 
\floatstyle{ruled}
\newfloat{listing}{tb}{lst}{}
\floatname{listing}{Listing}
\title{Graph-Based Cross-Domain Knowledge Distillation for Cross-Dataset Text-to-Image Person Retrieval}
\author{
    Bingjun Luo,
    Jinpeng Wang,
    Zewen Wang,
    Junjie Zhu\footnote{Corresponding author.},
    Xibin Zhao,
}
\affiliations{
    BNRist, KLISS, and School of Software, Tsinghua University\\
    luobingjun@gmail.com, \{wjp21,wang-zw19,zhujj18\}@mails.tsinghua.edu.cn, zxb@tsinghua.edu.cn
}

\usepackage{bibentry}

\usepackage{bm}
\usepackage{multirow}
\usepackage{subfigure}
\usepackage{booktabs} 
\usepackage{amsmath}
\usepackage{amssymb}

\begin{document}


\newcommand{\figleft}{{\em (Left)}}
\newcommand{\figcenter}{{\em (Center)}}
\newcommand{\figright}{{\em (Right)}}
\newcommand{\figtop}{{\em (Top)}}
\newcommand{\figbottom}{{\em (Bottom)}}
\newcommand{\captiona}{{\em (a)}}
\newcommand{\captionb}{{\em (b)}}
\newcommand{\captionc}{{\em (c)}}
\newcommand{\captiond}{{\em (d)}}

\newcommand{\newterm}[1]{{\bf #1}}

\def\figref#1{figure~\ref{#1}}
\def\Figref#1{Figure~\ref{#1}}
\def\twofigref#1#2{figures \ref{#1} and \ref{#2}}
\def\quadfigref#1#2#3#4{figures \ref{#1}, \ref{#2}, \ref{#3} and \ref{#4}}
\def\secref#1{section~\ref{#1}}
\def\Secref#1{Section~\ref{#1}}
\def\twosecrefs#1#2{sections \ref{#1} and \ref{#2}}
\def\secrefs#1#2#3{sections \ref{#1}, \ref{#2} and \ref{#3}}
\def\eqref#1{equation~\ref{#1}}
\def\Eqref#1{Equation~\ref{#1}}
\def\plaineqref#1{\ref{#1}}
\def\chapref#1{chapter~\ref{#1}}
\def\Chapref#1{Chapter~\ref{#1}}
\def\rangechapref#1#2{chapters\ref{#1}--\ref{#2}}
\def\algref#1{algorithm~\ref{#1}}
\def\Algref#1{Algorithm~\ref{#1}}
\def\twoalgref#1#2{algorithms \ref{#1} and \ref{#2}}
\def\Twoalgref#1#2{Algorithms \ref{#1} and \ref{#2}}
\def\partref#1{part~\ref{#1}}
\def\Partref#1{Part~\ref{#1}}
\def\twopartref#1#2{parts \ref{#1} and \ref{#2}}

\def\ceil#1{\lceil #1 \rceil}
\def\floor#1{\lfloor #1 \rfloor}
\def\1{\bm{1}}
\newcommand{\train}{\mathcal{D}}
\newcommand{\valid}{\mathcal{D_{\mathrm{valid}}}}
\newcommand{\test}{\mathcal{D_{\mathrm{test}}}}

\def\eps{{\epsilon}}

\def\reta{{\textnormal{$\eta$}}}
\def\ra{{\textnormal{a}}}
\def\rb{{\textnormal{b}}}
\def\rc{{\textnormal{c}}}
\def\rd{{\textnormal{d}}}
\def\re{{\textnormal{e}}}
\def\rf{{\textnormal{f}}}
\def\rg{{\textnormal{g}}}
\def\rh{{\textnormal{h}}}
\def\ri{{\textnormal{i}}}
\def\rj{{\textnormal{j}}}
\def\rk{{\textnormal{k}}}
\def\rl{{\textnormal{l}}}
\def\rn{{\textnormal{n}}}
\def\ro{{\textnormal{o}}}
\def\rp{{\textnormal{p}}}
\def\rq{{\textnormal{q}}}
\def\rr{{\textnormal{r}}}
\def\rs{{\textnormal{s}}}
\def\rt{{\textnormal{t}}}
\def\ru{{\textnormal{u}}}
\def\rv{{\textnormal{v}}}
\def\rw{{\textnormal{w}}}
\def\rx{{\textnormal{x}}}
\def\ry{{\textnormal{y}}}
\def\rz{{\textnormal{z}}}

\def\rvepsilon{{\mathbf{\epsilon}}}
\def\rvtheta{{\mathbf{\theta}}}
\def\rva{{\mathbf{a}}}
\def\rvb{{\mathbf{b}}}
\def\rvc{{\mathbf{c}}}
\def\rvd{{\mathbf{d}}}
\def\rve{{\mathbf{e}}}
\def\rvf{{\mathbf{f}}}
\def\rvg{{\mathbf{g}}}
\def\rvh{{\mathbf{h}}}
\def\rvu{{\mathbf{i}}}
\def\rvj{{\mathbf{j}}}
\def\rvk{{\mathbf{k}}}
\def\rvl{{\mathbf{l}}}
\def\rvm{{\mathbf{m}}}
\def\rvn{{\mathbf{n}}}
\def\rvo{{\mathbf{o}}}
\def\rvp{{\mathbf{p}}}
\def\rvq{{\mathbf{q}}}
\def\rvr{{\mathbf{r}}}
\def\rvs{{\mathbf{s}}}
\def\rvt{{\mathbf{t}}}
\def\rvu{{\mathbf{u}}}
\def\rvv{{\mathbf{v}}}
\def\rvw{{\mathbf{w}}}
\def\rvx{{\mathbf{x}}}
\def\rvy{{\mathbf{y}}}
\def\rvz{{\mathbf{z}}}

\def\erva{{\textnormal{a}}}
\def\ervb{{\textnormal{b}}}
\def\ervc{{\textnormal{c}}}
\def\ervd{{\textnormal{d}}}
\def\erve{{\textnormal{e}}}
\def\ervf{{\textnormal{f}}}
\def\ervg{{\textnormal{g}}}
\def\ervh{{\textnormal{h}}}
\def\ervi{{\textnormal{i}}}
\def\ervj{{\textnormal{j}}}
\def\ervk{{\textnormal{k}}}
\def\ervl{{\textnormal{l}}}
\def\ervm{{\textnormal{m}}}
\def\ervn{{\textnormal{n}}}
\def\ervo{{\textnormal{o}}}
\def\ervp{{\textnormal{p}}}
\def\ervq{{\textnormal{q}}}
\def\ervr{{\textnormal{r}}}
\def\ervs{{\textnormal{s}}}
\def\ervt{{\textnormal{t}}}
\def\ervu{{\textnormal{u}}}
\def\ervv{{\textnormal{v}}}
\def\ervw{{\textnormal{w}}}
\def\ervx{{\textnormal{x}}}
\def\ervy{{\textnormal{y}}}
\def\ervz{{\textnormal{z}}}

\def\rmA{{\mathbf{A}}}
\def\rmB{{\mathbf{B}}}
\def\rmC{{\mathbf{C}}}
\def\rmD{{\mathbf{D}}}
\def\rmE{{\mathbf{E}}}
\def\rmF{{\mathbf{F}}}
\def\rmG{{\mathbf{G}}}
\def\rmH{{\mathbf{H}}}
\def\rmI{{\mathbf{I}}}
\def\rmJ{{\mathbf{J}}}
\def\rmK{{\mathbf{K}}}
\def\rmL{{\mathbf{L}}}
\def\rmM{{\mathbf{M}}}
\def\rmN{{\mathbf{N}}}
\def\rmO{{\mathbf{O}}}
\def\rmP{{\mathbf{P}}}
\def\rmQ{{\mathbf{Q}}}
\def\rmR{{\mathbf{R}}}
\def\rmS{{\mathbf{S}}}
\def\rmT{{\mathbf{T}}}
\def\rmU{{\mathbf{U}}}
\def\rmV{{\mathbf{V}}}
\def\rmW{{\mathbf{W}}}
\def\rmX{{\mathbf{X}}}
\def\rmY{{\mathbf{Y}}}
\def\rmZ{{\mathbf{Z}}}

\def\ermA{{\textnormal{A}}}
\def\ermB{{\textnormal{B}}}
\def\ermC{{\textnormal{C}}}
\def\ermD{{\textnormal{D}}}
\def\ermE{{\textnormal{E}}}
\def\ermF{{\textnormal{F}}}
\def\ermG{{\textnormal{G}}}
\def\ermH{{\textnormal{H}}}
\def\ermI{{\textnormal{I}}}
\def\ermJ{{\textnormal{J}}}
\def\ermK{{\textnormal{K}}}
\def\ermL{{\textnormal{L}}}
\def\ermM{{\textnormal{M}}}
\def\ermN{{\textnormal{N}}}
\def\ermO{{\textnormal{O}}}
\def\ermP{{\textnormal{P}}}
\def\ermQ{{\textnormal{Q}}}
\def\ermR{{\textnormal{R}}}
\def\ermS{{\textnormal{S}}}
\def\ermT{{\textnormal{T}}}
\def\ermU{{\textnormal{U}}}
\def\ermV{{\textnormal{V}}}
\def\ermW{{\textnormal{W}}}
\def\ermX{{\textnormal{X}}}
\def\ermY{{\textnormal{Y}}}
\def\ermZ{{\textnormal{Z}}}

\def\vzero{{\bm{0}}}
\def\vone{{\bm{1}}}
\def\vmu{{\bm{\mu}}}
\def\vtheta{{\bm{\theta}}}
\def\vphi{{\bm{\phi}}}
\def\vsigma{{\bm{\sigma}}}
\def\va{{\bm{a}}}
\def\vb{{\bm{b}}}
\def\vc{{\bm{c}}}
\def\vd{{\bm{d}}}
\def\ve{{\bm{e}}}
\def\vf{{\bm{f}}}
\def\vg{{\bm{g}}}
\def\vh{{\bm{h}}}
\def\vi{{\bm{i}}}
\def\vj{{\bm{j}}}
\def\vk{{\bm{k}}}
\def\vl{{\bm{l}}}
\def\vm{{\bm{m}}}
\def\vn{{\bm{n}}}
\def\vo{{\bm{o}}}
\def\vp{{\bm{p}}}
\def\vq{{\bm{q}}}
\def\vr{{\bm{r}}}
\def\vs{{\bm{s}}}
\def\vt{{\bm{t}}}
\def\vu{{\bm{u}}}
\def\vv{{\bm{v}}}
\def\vw{{\bm{w}}}
\def\vx{{\bm{x}}}
\def\vy{{\bm{y}}}
\def\vz{{\bm{z}}}

\def\evalpha{{\alpha}}
\def\evbeta{{\beta}}
\def\evepsilon{{\epsilon}}
\def\evlambda{{\lambda}}
\def\evomega{{\omega}}
\def\evmu{{\mu}}
\def\evpsi{{\psi}}
\def\evsigma{{\sigma}}
\def\evtheta{{\theta}}
\def\eva{{a}}
\def\evb{{b}}
\def\evc{{c}}
\def\evd{{d}}
\def\eve{{e}}
\def\evf{{f}}
\def\evg{{g}}
\def\evh{{h}}
\def\evi{{i}}
\def\evj{{j}}
\def\evk{{k}}
\def\evl{{l}}
\def\evm{{m}}
\def\evn{{n}}
\def\evo{{o}}
\def\evp{{p}}
\def\evq{{q}}
\def\evr{{r}}
\def\evs{{s}}
\def\evt{{t}}
\def\evu{{u}}
\def\evv{{v}}
\def\evw{{w}}
\def\evx{{x}}
\def\evy{{y}}
\def\evz{{z}}

\def\mA{{\bm{A}}}
\def\mB{{\bm{B}}}
\def\mC{{\bm{C}}}
\def\mD{{\bm{D}}}
\def\mE{{\bm{E}}}
\def\mF{{\bm{F}}}
\def\mG{{\bm{G}}}
\def\mH{{\bm{H}}}
\def\mI{{\bm{I}}}
\def\mJ{{\bm{J}}}
\def\mK{{\bm{K}}}
\def\mL{{\bm{L}}}
\def\mM{{\bm{M}}}
\def\mN{{\bm{N}}}
\def\mO{{\bm{O}}}
\def\mP{{\bm{P}}}
\def\mQ{{\bm{Q}}}
\def\mR{{\bm{R}}}
\def\mS{{\bm{S}}}
\def\mT{{\bm{T}}}
\def\mU{{\bm{U}}}
\def\mV{{\bm{V}}}
\def\mW{{\bm{W}}}
\def\mX{{\bm{X}}}
\def\mY{{\bm{Y}}}
\def\mZ{{\bm{Z}}}
\def\mBeta{{\bm{\beta}}}
\def\mPhi{{\bm{\Phi}}}
\def\mLambda{{\bm{\Lambda}}}
\def\mSigma{{\bm{\Sigma}}}
\def\mTheta{{\bm{\Theta}}}
\def\mZero{{\bm{0}}}

\newcommand{\tens}[1]{\bm{\mathsfit{#1}}}
\def\tA{{\tens{A}}}
\def\tB{{\tens{B}}}
\def\tC{{\tens{C}}}
\def\tD{{\tens{D}}}
\def\tE{{\tens{E}}}
\def\tF{{\tens{F}}}
\def\tG{{\tens{G}}}
\def\tH{{\tens{H}}}
\def\tI{{\tens{I}}}
\def\tJ{{\tens{J}}}
\def\tK{{\tens{K}}}
\def\tL{{\tens{L}}}
\def\tM{{\tens{M}}}
\def\tN{{\tens{N}}}
\def\tO{{\tens{O}}}
\def\tP{{\tens{P}}}
\def\tQ{{\tens{Q}}}
\def\tR{{\tens{R}}}
\def\tS{{\tens{S}}}
\def\tT{{\tens{T}}}
\def\tU{{\tens{U}}}
\def\tV{{\tens{V}}}
\def\tW{{\tens{W}}}
\def\tX{{\tens{X}}}
\def\tY{{\tens{Y}}}
\def\tZ{{\tens{Z}}}

\def\gA{{\mathcal{A}}}
\def\gB{{\mathcal{B}}}
\def\gC{{\mathcal{C}}}
\def\gD{{\mathcal{D}}}
\def\gE{{\mathcal{E}}}
\def\gF{{\mathcal{F}}}
\def\gG{{\mathcal{G}}}
\def\gH{{\mathcal{H}}}
\def\gI{{\mathcal{I}}}
\def\gJ{{\mathcal{J}}}
\def\gK{{\mathcal{K}}}
\def\gL{{\mathcal{L}}}
\def\gM{{\mathcal{M}}}
\def\gN{{\mathcal{N}}}
\def\gO{{\mathcal{O}}}
\def\gP{{\mathcal{P}}}
\def\gQ{{\mathcal{Q}}}
\def\gR{{\mathcal{R}}}
\def\gS{{\mathcal{S}}}
\def\gT{{\mathcal{T}}}
\def\gU{{\mathcal{U}}}
\def\gV{{\mathcal{V}}}
\def\gW{{\mathcal{W}}}
\def\gX{{\mathcal{X}}}
\def\gY{{\mathcal{Y}}}
\def\gZ{{\mathcal{Z}}}

\def\sA{{\mathbb{A}}}
\def\sB{{\mathbb{B}}}
\def\sC{{\mathbb{C}}}
\def\sD{{\mathbb{D}}}
\def\sF{{\mathbb{F}}}
\def\sG{{\mathbb{G}}}
\def\sH{{\mathbb{H}}}
\def\sI{{\mathbb{I}}}
\def\sJ{{\mathbb{J}}}
\def\sK{{\mathbb{K}}}
\def\sL{{\mathbb{L}}}
\def\sM{{\mathbb{M}}}
\def\sN{{\mathbb{N}}}
\def\sO{{\mathbb{O}}}
\def\sP{{\mathbb{P}}}
\def\sQ{{\mathbb{Q}}}
\def\sR{{\mathbb{R}}}
\def\sS{{\mathbb{S}}}
\def\sT{{\mathbb{T}}}
\def\sU{{\mathbb{U}}}
\def\sV{{\mathbb{V}}}
\def\sW{{\mathbb{W}}}
\def\sX{{\mathbb{X}}}
\def\sY{{\mathbb{Y}}}
\def\sZ{{\mathbb{Z}}}

\def\emLambda{{\Lambda}}
\def\emA{{A}}
\def\emB{{B}}
\def\emC{{C}}
\def\emD{{D}}
\def\emE{{E}}
\def\emF{{F}}
\def\emG{{G}}
\def\emH{{H}}
\def\emI{{I}}
\def\emJ{{J}}
\def\emK{{K}}
\def\emL{{L}}
\def\emM{{M}}
\def\emN{{N}}
\def\emO{{O}}
\def\emP{{P}}
\def\emQ{{Q}}
\def\emR{{R}}
\def\emS{{S}}
\def\emT{{T}}
\def\emU{{U}}
\def\emV{{V}}
\def\emW{{W}}
\def\emX{{X}}
\def\emY{{Y}}
\def\emZ{{Z}}
\def\emSigma{{\Sigma}}

\newcommand{\etens}[1]{\mathsfit{#1}}
\def\etLambda{{\etens{\Lambda}}}
\def\etA{{\etens{A}}}
\def\etB{{\etens{B}}}
\def\etC{{\etens{C}}}
\def\etD{{\etens{D}}}
\def\etE{{\etens{E}}}
\def\etF{{\etens{F}}}
\def\etG{{\etens{G}}}
\def\etH{{\etens{H}}}
\def\etI{{\etens{I}}}
\def\etJ{{\etens{J}}}
\def\etK{{\etens{K}}}
\def\etL{{\etens{L}}}
\def\etM{{\etens{M}}}
\def\etN{{\etens{N}}}
\def\etO{{\etens{O}}}
\def\etP{{\etens{P}}}
\def\etQ{{\etens{Q}}}
\def\etR{{\etens{R}}}
\def\etS{{\etens{S}}}
\def\etT{{\etens{T}}}
\def\etU{{\etens{U}}}
\def\etV{{\etens{V}}}
\def\etW{{\etens{W}}}
\def\etX{{\etens{X}}}
\def\etY{{\etens{Y}}}
\def\etZ{{\etens{Z}}}

\newcommand{\pdata}{p_{\rm{data}}}
\newcommand{\ptrain}{\hat{p}_{\rm{data}}}
\newcommand{\Ptrain}{\hat{P}_{\rm{data}}}
\newcommand{\pmodel}{p_{\rm{model}}}
\newcommand{\Pmodel}{P_{\rm{model}}}
\newcommand{\ptildemodel}{\tilde{p}_{\rm{model}}}
\newcommand{\pencode}{p_{\rm{encoder}}}
\newcommand{\pdecode}{p_{\rm{decoder}}}
\newcommand{\precons}{p_{\rm{reconstruct}}}

\newcommand{\E}{\mathbb{E}}
\newcommand{\Ls}{\mathcal{L}}
\newcommand{\R}{\mathbb{R}}
\newcommand{\emp}{\tilde{p}}
\newcommand{\lr}{\alpha}
\newcommand{\reg}{\lambda}
\newcommand{\rect}{\mathrm{rectifier}}
\newcommand{\softmax}{\mathrm{softmax}}
\newcommand{\sigmoid}{\sigma}
\newcommand{\softplus}{\zeta}
\newcommand{\KL}{D_{\mathrm{KL}}}
\newcommand{\Var}{\mathrm{Var}}
\newcommand{\standarderror}{\mathrm{SE}}
\newcommand{\Cov}{\mathrm{Cov}}
\newcommand{\normlzero}{L^0}
\newcommand{\normlone}{L^1}
\newcommand{\normltwo}{L^2}
\newcommand{\normlp}{L^p}
\newcommand{\normmax}{L^\infty}

\newcommand{\parents}{Pa} 

\let\ab\allowbreak

\maketitle

\begin{abstract}
Video surveillance systems are crucial components for ensuring public safety and management in smart city.
As a fundamental task in video surveillance, text-to-image person retrieval aims to retrieve the target person from an image gallery that best matches the given text description. 
Most existing text-to-image person retrieval methods are trained in a supervised manner that requires sufficient labeled data in the target domain. 
However, it is common in practice that only unlabeled data is available in the target domain due to the difficulty and cost of data annotation, which limits the generalization of existing methods in practical application scenarios.
To address this issue, we propose a novel unsupervised domain adaptation method, termed Graph-Based Cross-Domain Knowledge Distillation (GCKD), to learn the cross-modal feature representation for text-to-image person retrieval in a cross-dataset scenario.
The proposed GCKD method consists of two main components.
Firstly, a graph-based multi-modal propagation module is designed to bridge the cross-domain correlation among the visual and textual samples.
Secondly, a contrastive momentum knowledge distillation module is proposed to learn the cross-modal feature representation using the online knowledge distillation strategy.
By jointly optimizing the two modules, the proposed method is able to achieve efficient performance for cross-dataset text-to-image person retrieval. 
Extensive experiments on three publicly available text-to-image person retrieval datasets demonstrate the effectiveness of the proposed GCKD method, which consistently outperforms the state-of-the-art baselines.

\end{abstract}

%

\section{Introduction}

\begin{figure}[t]

    \includegraphics[width=1\linewidth]{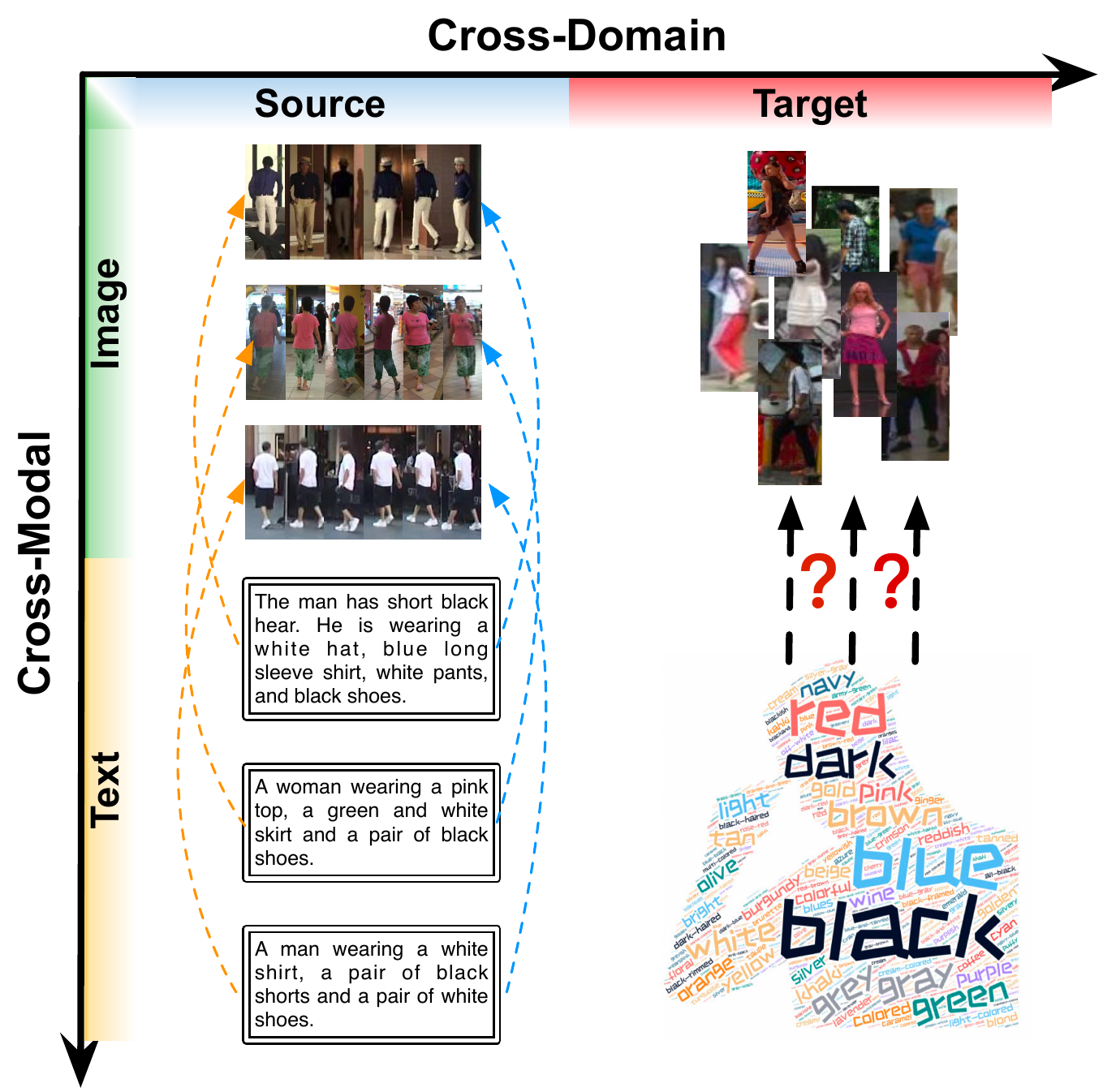}
    
    \caption{An illustration of cross-dataset text-to-image person retrieval task. The source dataset contains paired image-text annotations, while the target dataset only contains unpaired images and texts. This task faces both cross-domain shift and cross-modality gap challenges. 
    Part of the figures comes from \cite{ding2021semantically, zhu2021dssl, zhu2024improving}. 
    }
    \label{fig1}
\end{figure}

Person retrieval is a fundamental problem in the field of video surveillance, which has attracted extensive attention from both scientific research and practical applications \cite{jing2020cross, li2023multi}.
Text-to-image person retrieval is a subtask of person retrieval, which aims to retrieve the target person from the image gallery that matches the given text description \cite{li2017person, yan2023clip}. This task is associated with both image-text retrieval \cite{li2021align} and image-based person retrieval \cite{he2021transreid}.
Different from image-based person retrieval, text-to-image person retrieval is able to retrieve the target person only based on the given query text, which is more user-friendly and easily accessible than using query image \cite{jiang2023cross, bai2023rasa}. 
Thus, text-to-image person retrieval has attracted increasing attention in recent years \cite{zhu2021dssl, shu2022see, zhu2024improving}. 
With the advancement of deep learning technology \cite{fan2024instantsplat, gu2020cascade}, the performance of text-to-image person retrieval has significantly improved. Recently, large language models \cite{radford2021learning, li2021align, shao2024stllm} have further enhanced this performance.

However, most existing text-to-image person retrieval methods are trained in the supervised manner, which requires a large amount of high-quality annotated image-text alignment datasets in the target application scenario.
These methods usually rely on large-scale alignment data collection and manual annotation, which are difficult to obtain and expensive. This limits the generalization of these methods in practical applications \cite{jiang2023cross, yan2023clip}.

A feasible solution is to adopt the transfer learning paradigm, which transfers the knowledge learned from the existing large-scale supervised datasets to the target unlabeled dataset, i.e., cross-dataset text-to-image person retrieval as shown in Fig. \ref{fig1}.
In the cross-dataset text-to-image person retrieval task, there are mainly two challenges: 
\textbf{1) Domain shift.} For the cross-domain transferring task, an important challenge is the domain shift, i.e., the distribution difference between the source and target datasets.
\textbf{2) Modality gap.} Another important challenge for this task is the modality heterogeneity between visual and textual modalities \cite{jiang2023cross}.
In the visual modality, data is usually high-dimensional and continuous, where differences mainly lie in the background, pose, lighting, etc. In contrast, in the textual modality, data is usually low-dimensional and discrete, where differences mainly lie in semantic ambiguity and order.
In the cross-dataset scenario, due to the lack of text-image matching annotations on the target dataset, the modality heterogeneity will be further exacerbated.
In the cross-dataset text-to-image person retrieval task, the superposition of these two challenges makes the performance of existing text-to-image person retrieval methods decrease in cross-domain scenarios, making this task more difficult.

To address the above challenges, we propose a novel unsupervised domain adaptation method for cross-dataset text-to-image person retrieval, named \textbf{Graph-based Cross-domain Knowledge Distillation (GCKD)}. The proposed method is based on the vision language pre-training models, which have shown great power in the field of text-to-image person retrieval recently \cite{yan2023clip, bai2023rasa, jiang2023cross}.
The proposed GCKD method consists of two novel components. The first component is a graph-based multi-domain propagation (GMP) module, 
which aims to address the domain shift problem by propagating feature information between the source and target domains on the dynamic cross-domain graph.
The second component is a contrastive momentum knowledge distillation (CMKD) module,
which aims to address the modality gap by constructing high-confidence pseudo text-image similarity labels through momentum knowledge distillation, which can guide the model to learn modal-invariant feature representation.
By integrating these two modules, our method can effectively address the domain shift and modality gap challenges, and achieve more accurate and robust performance in the cross-dataset text-to-image person retrieval task.

In summary, the contributions of this paper are three-fold:
\begin{itemize}
    \item We propose a novel unsupervised domain adaptation method, i.e. GCKD, to improve the cross-domain performance of text-to-image person retrieval. To the best of our knowledge, this is the first work to adopt the vision language pre-training models for cross-dataset text-to-image person retrieval.
    \item We introduce a cross-domain graph propagation mechanism to address the domain shift challenge, and a contrastive momentum knowledge distillation strategy to address the modality gap problem. These components are integrated into a unified framework to learn the cross-modal representation.
    \item We conduct extensive experiments on three commonly used datasets, and the results demonstrate that our proposed method outperforms the state-of-the-art methods in the cross-dataset text-to-image person retrieval task.
\end{itemize}

\section{Related Work}
\subsection{Text-to-Image Person Retrieval}
With the advancement of smart city \cite{xi2024evaluating, xi2024optimizing}, the integration of text-to-image person retrieval is gaining increasing attention.
The primary challenge in text-to-image person retrieval is cross-modal alignment, which can be categorized into two main strategies: cross-modal interaction-based and cross-modal interaction-free methods.   
Cross-modal interaction-based methods~\cite{li2017person,zheng2020gumbel,zhu2021dssl} utilize attention mechanisms to identify local correspondences (e.g., patch-word, patch-phrase) between images and texts, predicting matching scores for image-text pairs. Notably, DSSL~\cite{zhu2021dssl} separates visual data into person and surroundings information for effective surroundings-person distinction.  
In contrast, cross-modal interaction-free methods~\cite{chen2021cross,bai2023rasa,shu2022see} achieve high performance without complex interactions. The success of Transformer architectures in Vision and Language Tasks has led to the development of various Transformer-based models~\cite{li2022learning,shao2022learning}, such as LGUR~\cite{shao2022learning}, which learns granularity-unified representations for text and image modalities in an end-to-end manner.  

While existing methods typically require labeled data for downstream tasks, our work operates in an unsupervised setting, eliminating the need for extensive alignment data collection and manual annotation, thus demonstrating greater efficiency in cross-domain scenarios.

\subsection{Unsupervised Domain Adaptation}
Unsupervised domain adaptation (UDA) seeks to transfer knowledge from a labeled source domain to an unlabeled target domain, but its application in text-to-image person retrieval is limited. MAN~\cite{jing2020cross} introduces a moment alignment network for cross-modal text-image alignment, but its reliance on CNN and LSTM limits adaptability to Transformer architectures.   
POUF~\cite{tanwisuth2023pouf} aligns prototypes and target data in latent space using transport-based distribution alignment and mutual information maximization. ReCLIP~\cite{hu2024reclip} presents a source-free domain adaptation method for vision-language models through cross-modality self-training with learned pseudo labels. However, both POUF and ReCLIP face modality heterogeneity issues specific to text-to-image person retrieval tasks.  

In contrast to these approaches, our model leverages vision-language pre-training for unsupervised domain adaptation in text-to-image person retrieval, effectively addressing challenges related to domain shift and modality gaps.
\begin{figure*}[htbp]
    \includegraphics[width=\linewidth]{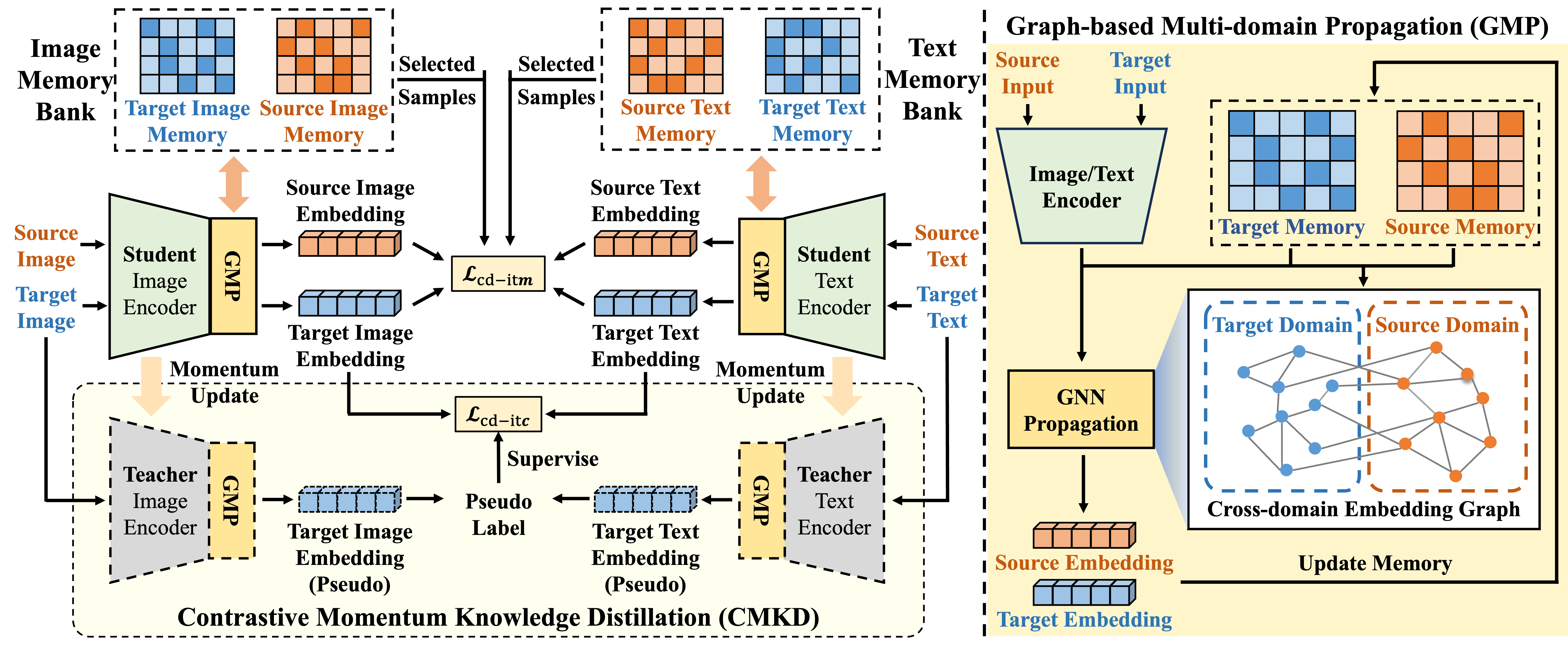}
    \caption{The main framework of the proposed Graph-based Cross-domain Knowledge Distillation (GCKD) method. The proposed method consists of two main components: Graph-based Multi-domain Propagation (GMP) module and Contrastive Momentum Knowledge Distillation (CMKD) module.}
    \label{fig:framework}
\end{figure*}

\section{Problem Definition}
Denote $\gD_s=\{\gV_s, \gT_s\}$ as a well-annotated text-to-image person retrieval dataset from the source domain. The source dataset $\gD_s$ consists of a labeled image set $\gV_s=\{(\vv_s^{(i)}, y_s^{(i)})\}_{i=1}^{M_s}$ of size $M_s$ and a labeled text set $\gT_s=\{(\vt_s^{(j)}, y_s^{(j)})\}_{j=1}^{N_s}$ of size $N_s$ respectively, where $\vv_s^{(i)}$ and $y_s^{(i)}$ are the image and corresponding identity label of the $i$-th visual sample, and $\vt_s^{(j)}$ and $y_s^{(j)}$ are the text and corresponding identity label of the $j$-th textual sample.
Denote $\gD_t=\{\gV_t, \gT_t\}$ as an unlabeled dataset from the target domain.
The target dataset $\gD_t$ only contains unpaired images $\gV_t=\{\vv_t^{(i)}\}_{i=1}^{M_t}$ and texts $\gT_t=\{\vt_t^{(j)}\}_{j=1}^{N_t}$, where $\vv_t^{(i)}$ and $\vt_t^{(j)}$ are the $i$-th image and $j$-th text sample in the target dataset, respectively.
Given the labeled source dataset $\gD_s$ and the unlabeled target dataset $\gD_t$, the goal of cross-dataset text-to-image person retrieval is to learn a model that can effectively retrieve the target person image $\vv_t$ for each given text query $\vt_t$ in the target domain.

\section{Our Model}
In this section, we present the proposed Graph-based Cross-domain Knowledge Distillation (GCKD) method for cross-dataset text-to-image person retrieval.

\subsection{Model Overview}
In this paper, we propose a novel graph-based cross-dataset text-to-image person retrieval method, named \textbf{Graph-based Cross-domain Knowledge Distillation (GCKD)}. As shown in Fig. \ref{fig:framework}, the proposed GCKD method consists of two main components as the following.
\textbf{Firstly}, a graph-based multi-domain propagation module is proposed to address the domain shift problem in the cross-modal retrieval task. By unifying visual and textual features from the source and target domains into a holistic cross-domain embedding graph, the module is designed to bridge the correlation and align the cross-modal features between the source and target domain.
\textbf{Secondly}, a contrastive momentum knowledge distillation module is proposed to address the modality gap problem in the cross-domain scenario. Different from the single-domain visual language model, the module introduces momentum strategy into the domain adaptation task and proposes cross-domain fine-grained matching tasks to learn the shared representation among different modalities and domains.
By jointly optimizing the two modules, the proposed model is able to achieve efficient performance for cross-dataset text-to-image person retrieval.

\subsection{Graph-based Multi-domain Propagation}
The domain shift problem, arising from variations in image conditions (e.g., resolutions, angles) and text styles (e.g., descriptive approaches, paragraph lengths) across datasets, significantly impacts the performance of text-to-image person retrieval models. To mitigate this, we propose a graph-based multi-domain propagation module that connects features from both source and target domains. By utilizing a unified graph, this module bridges correlations and reduces discrepancies across domains. 

\subsubsection{Embedding Memory}\label{sec:feature-storing}
The embedding memory stage is designed to store the embeddings from the source and target domains.
We propose two types of memory banks to store the most recent $C$ embedding of $D$-dimmension from the visual and textual modalities during training. The image memory bank is further composed of the source image memory $Q_{SI}\in\sR^{C\times D}$ and $Q_{TI}\in\sR^{C\times D}$, which stores the image embeddings of source and target domains respectively.
The setting is similar for the textual modal, which stores the most recent $C$ text embeddings of source and target domains respectively, i.e., $Q_{ST}\in\sR^{C\times D}$ and $Q_{TT}\in\sR^{C\times D}$.
At each training iteration, the image and text memories are updated iteratively from the recent embeddings of both source and target domains.
By leveraging the memory banks and the corresponding updating mechanism, the module can capture the multi-modal embedding information across batches and even epochs, which effectively extends the scope of multi-domain propagation.

\subsubsection{Graph Construction}
For each training iteration, we are given a batch of image/text embeddings $\mF_m=(\vf_m^{(1)}, \cdots, \vf_m^{(K)})\in\sR^{B\times K}$ from target domains, where $m=I$ denotes image modality embeddings and $m=T$ denotes text modality embeddings. Based on the input embeddings and the existing source and target memories, the dynamic cross-domain graph $\gG$ is constructed using the $K$-nearest neighbor (KNN) algorithm.

Specifically, the graph $\gG$ can be determined by the vertex set $\gV$, the vertex embedding matrix $\mX$, and the adjacency matrix $\mA$. The vertex of the graph comes from three parts:
\begin{equation}
    \gV = \gV_\text{input} \cup \gV_\text{src\_mem} \cup \gV_\text{tgt\_mem}
\end{equation}
where $\gV_\text{input}=\{v_i\}_{i=1}^B$ is the vertex set of the given image/text batch $\mF_m$, $\gV_\text{src\_mem}=\{v_i\}_{i=B+1}^{B+C}$ and $\gV_\text{tgt\_mem}=\{v_i\}_{i=B+C+1}^{B+2C}$ are the vertex set of the source and target memories $Q_{Sm}$, $Q_{Tm}$, respectively. The vertex embedding matrix $\mX\in\sR^{(B+2C)\times K}$ is constructed by concatenating the input embeddings and the embeddings from the source and target memories:
\begin{equation}
    \mX = 
    (\mF_m^\top, Q_{Sm}^\top, Q_{Tm}^\top)^\top
\end{equation}
The adjacency matrix $\mA\in\sR^{(B+2C)\times(B+2C)}$ is dynamically computed by the KNN algorithm based on the cosine similarity in vertex embedding matrix $\mX$:
\begin{equation}
    \mA_{ij} = \begin{cases}
        1, & \text{if } v_i \text{ is one of the } K \text{ nearest neighbors of } v_j \\
        0, & \text{otherwise}
    \end{cases}
\end{equation}
By constructing a dynamic cross-domain graph $\gG$ from the input embeddings and source and target memories, the module maps multi-domain embeddings into a unified graph structure, bridging multi-domain correlations.

\subsubsection{GNN Propagation}
After constructing the dynamic cross-domain graph $\gG$, we introduce a Graph Neural Network (GNN) to propagate embeddings across domains, effectively learning the graph structure and capturing sample correlations. Specifically, we adopt a two-layer GNN model for this propagation.
The GNN propagation is performed by iteratively updating the vertex embedding matrix $\mX$ based on the adjacency matrix $\mA$ and the vertex embedding matrix $\mX$. For the $l$-th layer of the GNN, the update rule can be formulated as:
\begin{equation}
    \mX^{(l+1)} = \text{GNNConv}(\mX^{(l)}, \mA; \mTheta^{(l)})
\end{equation}
where $\text{GNNConv}(\cdot)$ is the GNN convolution operation, and $\mTheta^{(l)}$ is the learnable parameters of the $l$-th layer.

By iteratively updating the vertex embedding matrix $\mX$ for $L=2$ layers, the last layer is used to generate the final vertex embedding matrix $\mX^{(L)}$, which is further used to compute the final domain-aware embeddings $\overline{\mF}_m=\mX^{(L)}_{1:B}$ for the image/text samples in the target domain.
By propagating the embeddings across the dynamic cross-domain graph $\gG$, the module is able to bridge the correlation among the samples from both source and target domains and reduce the domain discrepancy in the cross-dataset scenario.

\subsection{Contrastive Momentum Knowledge Distillation}
The modality difference between text and image data creates a significant challenge known as the modality gap. Existing contrastive learning methods, primarily designed for single-domain visual-language tasks, often degrade in cross-domain scenarios \cite{li2021align}. To address this, we propose a novel contrastive momentum knowledge distillation module that combines cross-modal contrastive learning with cross-domain knowledge distillation. 

\subsubsection{Cross-domain Momentum Distillation}
As shown in Fig. \ref{fig:framework}, the backbone model consists of an image encoder and a text encoder, referred to as the student model $\text{En}_\text{Student}(\cdot, \mTheta_\text{Student})$ in knowledge distillation, which extracts visual and textual features from input samples. We also create a teacher model with the same structure, $\text{En}_\text{Teacher}(\cdot, \mTheta_\text{Teacher})$. Both models are initialized with pre-trained weights from the source dataset, which provide rich knowledge from the source domain.

During the training process, the parameters of the teacher model are not optimized by the gradient descent method, but are updated by the exponential moving average (EMA) mechanism as follows:
\begin{equation}
    \mTheta_\text{Teacher} \leftarrow m \mTheta_\text{Teacher} + (1-m) \mTheta_\text{Student}
\end{equation}
where $m\in[0,1]$ is the momentum coefficient. 
By updating the teacher model with EMA, its parameters lag behind those of the student model, retaining more source domain knowledge. This allows the teacher model to generate pseudo labels that encapsulate source domain knowledge, guiding the student model to learn better representations in the target domain.

\subsubsection{Cross-modal Contrastive Learning}
After generating pseudo labels from the teacher model, the next challenge is effective cross-domain contrastive learning for knowledge transfer and improved representations in the target domain. We propose a cross-modal image-text contrast loss that leverages source domain knowledge from the teacher model to enhance target domain representations.

Specifically, given a batch of paired source domain samples $(\vv_s, \vt_s)$, and unpaired target domain visual and textual samples $\vv_t$ and $\vt_t$, the student model is used to generate the pseudo target domain image features $\hat{\vf}_{TI}$ and pseudo target domain text features $\hat{\vf}_{TT}$.
At the same time, the student model is used to extract the source domain image features $\vf_{SI}$ and source domain text features $\vf_{ST}$, and the domain-aware target domain image features $\vf_{TI}$ and target domain text features $\vf_{TT}$.
The cross-domain image-text contrast loss is defined as:
\begin{equation}
    \begin{aligned}
    \gL_{cd-itc}=&-\sum_{\vf_{TI}, \vf_{TT}}s_\text{i2t}\log\frac{\exp(d(\vf_{TI}, \vf_{TT})/\tau)}{\sum_{q\in Q_{TT}}\exp(d(\vf_{TI}, q)/\tau)} \\
    &-\sum_{\vf_{TT}, \vf_{TI}}s_\text{t2i}\log\frac{\exp(d(\vf_{TT}, \vf_{TI})/\tau)}{\sum_{q\in Q_{TI}}\exp(d(\vf_{TT}, q)/\tau)}
    \end{aligned}
    \label{eq:cd-itc}
\end{equation}
where 
\begin{align}
    s_\text{i2t} & = \frac{\exp(d(\hat{\vf}_{TI}, \hat{\vf}_{TT})/\tau)}{\sum_{q_{TT}\in Q_{TT}}\exp(d(\hat{\vf}_{TI}, q_{TT})/\tau)} \\
    s_\text{t2i} & = \frac{\exp(d(\hat{\vf}_{TT}, \hat{\vf}_{TI})/\tau)}{\sum_{q_{TI}\in Q_{TI}}\exp(d(\hat{\vf}_{TT}, q_{TI})/\tau)}
\end{align}
are the pseudo similarity targets generated by the teacher model, $d(\cdot, \cdot)$ is the cosine similarity function, $\tau$ is the temperature parameter, $\mQ_{TT}, \mQ_{TI}$ are the target text and image memory respectively.
By introducing the memory banks, the contrastive loss function is extended to almost all samples in the target domain, enabling the model to explore more positive and negative samples and facilitating feature extraction.

\subsubsection{Cross-domain Fine-grained Matching}
As noted in \cite{li2021align}, the image-text matching task is a binary classification problem, with positive pairs from the same identity. To enhance fine-grained matching in the target domain, we propose a cross-domain image-text matching task using target-domain positive pairs and cross-domain hard negative pairs. Positive pairs are generated from high-confidence pseudo labels from the teacher model, while negative pairs consist of the most challenging cross-domain pairs for the student model to distinguish.

For each target domain image feature $\vf_{TI}$, the positive text feature $\vf_{TT}$ is selected from the target text memory $Q_{TT}$ only if $d(\vf_{TI},\vf_{TT}) > \delta$ where $\delta$ is a predefined threshold. The negative text feature $\vf_{ST}$ is selected from the source domain feature batch if $\vf_{TI}$ has the highest cosine similarity with $\vf_{ST}$, i.e., $\vf_{TI}$ and $\vf_{ST}$ comes from different identities (even different domains) but have the highest similarity score.
Given the image and text sample pair, the multimodal encoder in the backbone network is utilized to produce the binary matching probability output $\hat{p}(\cdot, \cdot)$ between the given pairs.
After a softmax function, the cross-domain image-text matching loss is defined as:
\begin{equation}
    \begin{aligned}
    \gL_{cd-itm} &= -\sum_{(\vf_{TI}, \vf_{TT})}\log \text{Softmax}(\hat{p}(\vf_{TI}, \vf_{TT})) \\
    &- \sum_{(\vf_{TI}, \vf_{ST})}\log(1-\text{Softmax}(\hat{p}(\vf_{TI}, \vf_{ST})))
    \end{aligned}
    \label{eq:cd-itm}
\end{equation}

The overall optimization objective is as follows:
\begin{equation}
\gL=\lambda_1 \gL_{cd-itc} + \lambda_2 \gL_{cd-itm} + \lambda_3 \gL_{mlm}
\end{equation}
where $\lambda_1, \lambda_2, \lambda_3$ are the coefficient hyperparameters for each loss term, $\gL_{cd-itc}$ is the cross-domain image-text contrast loss defined in Eq. (\ref{eq:cd-itc}), $\gL_{cd-itm}$ is the cross-domain fine-grained matching loss defined in Eq. (\ref{eq:cd-itm}), $\gL_{mlm}$ is the masked language model loss proposed by \cite{li2021align}.

\begin{table*}[h]
    \centering
\begin{tabular}{cc|c|cccc|cccc}
\hline
\multirow{2}[0]{*}{Train Set} & \multirow{2}[0]{*}{Method} & \multirow{2}[0]{*}{Ref} & \multicolumn{4}{c|}{ICFG-PEDES $\rightarrow$ RSTPReid} & \multicolumn{4}{c}{ICFG-PEDES $\rightarrow$ CUHK-PEDES} \\
\cline{4-11}      &       &       & Rank-1 & Rank-5 & Rank-10 & mAP   & Rank-1 & Rank-5 & Rank-10 & mAP \\
\hline
\multirow{3}[0]{*}{Supervised} & IRRA & CVPR'23 & 60.20  & 81.30  & 88.20 & 47.17  & 73.38  & 89.93  & 93.71  & 66.13  \\
      & APTM & MM'23 & 67.50  & 85.70  & 91.45  & 52.56  & 76.53  & 90.04  & 94.15  & 66.91  \\
      & RaSa & IJCAI'23 &       66.90 & 86.50  & 91.35  &       52.31 & 76.51  & 90.29   & 94.25  & 69.38  \\
\hline
\hline
\multirow{5}[0]{*}{Source Only} & IVT & ECCVW'22 & 43.70  & 65.10  & 75.55  & 37.65  & 22.63  & 42.29  & 52.36  & 19.85  \\
      & CFine & TIP'23 & 47.40  & 70.60  & 79.35  & 42.30  & 32.67  & 54.03  & 63.68  & 25.47  \\
      & IRRA & CVPR'23 & 45.10  & 69.20  & 78.75  & 36.76  & 33.43  & 56.11  & 66.23  & 31.38  \\
      & APTM & MM'23 & 52.50  & \underline{75.15}  & \underline{81.70}  & 40.81  & 46.44  & 66.89  & 74.45  & 40.14  \\
      & RaSa & IJCAI'23 & \underline{55.00}  & 73.65  & 81.55  & \underline{46.18}  & \underline{48.65}  & \underline{69.90}  & \underline{76.53}  & \underline{42.03}  \\
\hline
\multirow{3}[0]{*}{Source \& Target} & POUF  & ICML'23 & 36.60  & 61.80  & 73.15  & 28.28   & 20.50  & 39.39  & 49.03  & 18.48 \\
      & ReCLIP & WACV'24 & 50.35  & 73.25  & 81.20  & 42.75   & 33.48  & 55.25  & 64.26  & 29.19 \\
      & \textbf{Ours}  & /     & \textbf{59.95} & \textbf{79.05} & \textbf{85.95} & \textbf{49.68}  & \textbf{52.70} & \textbf{72.76} & \textbf{80.15} & \textbf{45.97}\\
\hline
\end{tabular}%
    \caption{Comparison results with state-of-the-art methods on ICFG-PEDES as source dataset. The performance in the supervised setting is reported for reference. The best results are emphasized in \textbf{bold}. The second-best results are noted by \underline{underline}.}
        \label{tab:comparison-icfg}%
\end{table*}%

\begin{table*}[h]
    \centering
\begin{tabular}{cc|c|cccc|cccc}
\hline
\multirow{2}[0]{*}{Train Set} & \multirow{2}[0]{*}{Method} & \multirow{2}[0]{*}{Ref} & \multicolumn{4}{c|}{RSTPReid $\rightarrow$ ICFG-PEDES} & \multicolumn{4}{c}{RSTPReid $\rightarrow$ CUHK-PEDES} \\
\cline{4-11}      &       &       & Rank-1 & Rank-5 & Rank-10 & mAP   & Rank-1 & Rank-5 & Rank-10 & mAP \\
\hline
\multirow{3}[0]{*}{Supervised} & IRRA & CVPR'23 & 63.46  & 80.25  & 85.82  & 38.06  & 73.38  & 89.93  & 93.71  & 66.13  \\
      & APTM & MM'23 & 68.51  & 82.99  & 87.56  & 41.22  & 76.53  & 90.04  & 94.15  & 66.91  \\
      & RaSa & IJCAI'23 & 65.28  & 80.40  & 85.12  & 41.29  & 76.51 & 90.29	 & 94.25  & 69.38 \\
\hline
\hline
\multirow{5}[0]{*}{Source Only} & IVT & ECCVW'22 & 19.43  & 35.22  & 43.83  & 13.52  & 16.73  & 33.65  & 43.01  & 14.42  \\
      & CFine & TIP'23 & 24.83  & 39.49  & 46.92  & 17.04  & 20.79  & 38.76  & 48.93  & 15.77  \\
      & IRRA & CVPR'23 & 32.35  & 49.68  & 57.74  & 20.57  & 32.65  & 55.20  & 65.38  & 30.17  \\
      & APTM & MM'23 & \underline{44.01}  & \underline{60.09}  & \underline{66.25}  & \underline{25.60}  & \underline{44.92}  & \underline{64.90}  & \underline{73.75}  & \underline{38.47}  \\
      & RaSa & IJCAI'23 & 41.30  & 56.18  & 62.36  & 22.39  & 42.85  & 61.97  & 69.79  & 35.64  \\
\hline
\multirow{3}[0]{*}{Source \& Target} & POUF & ICML'23 & 21.27  & 37.87  & 46.66  & 11.04  & 17.84  & 35.62  & 46.72  & 16.30  \\
      & ReCLIP & WACV'24 & 27.18  & 42.51  & 50.28  & 17.83  & 21.04  & 40.29  & 50.28  & 18.92  \\
      & \textbf{Ours} & /     & \textbf{46.40} & \textbf{61.17} & \textbf{66.97} & \textbf{26.06} & \textbf{51.01} & \textbf{70.29} & \textbf{77.29} & \textbf{43.71} \\
\hline
\end{tabular}%
    \caption{Comparison results with state-of-the-art methods on RSTPReid as source dataset. The performance in the supervised setting is reported for reference. The best results are emphasized in \textbf{bold}. The second-best results are noted by \underline{underline}.}
    \label{tab:comparison-rstp}%
\end{table*}%

\section{Experiment Setup}
In this section, we introduce the experiment setup of this paper, including datasets, baselines, task settings, evaluation metrics, and implementation details.

\subsection{Datasets}
In this paper, we conduct experiments on three publicly available text-to-image person retrieval datasets: ICFG-PEDES, RSTPReid, and CUHK-PEDES. 
\textbf{ICFG-PEDES} \cite{ding2021semantically} is the largest public dataset for text-to-image person retrieval, which consists of 54,522 images from 4,102 identities in total. 
\textbf{RSTPReid} \cite{zhu2021dssl} is a newly released text-to-image person retrieval dataset. The dataset comprises 20,505 images from 4,101 identities.
\textbf{CUHK-PEDES} \cite{li2017person} is also a commonly used dataset in the text-to-image person retrieval field, which is composed of 5 parts from different scenarios: SSM, VIPER, CUHK01, CUHK03, and Market-1501. 

\subsection{Baselines}\label{sec:baselines}
To comprehensively evaluate the proposed method, we follow recent works \cite{hao2023dual,zhu2024improving} to select state-of-the-art baselines and assess them in two training settings: Source Only (SO) and Source and Target (ST).

For the \textbf{Source Only (SO)} setting, we select several state-of-the-art methods from the single-domain text-to-image person retrieval task, trained solely on the labeled source dataset and tested directly in the target domain.
For the main experiments, the baselines include RaSa \cite{bai2023rasa}, APTM \cite{yang2023towards}, IRRA \cite{jiang2023cross}, CFine \cite{yan2023clip}, IVT \cite{shu2022see}. For the intra-dataset experiments on CUHK-PEDES, the baselines include EAIBC \cite{zhu2024improving}, RaSa \cite{bai2023rasa}, SSAN \cite{ding2021semantically}, MIA \cite{niu2020improving}, SCAN \cite{lee2018stacked}, CMPM-CMPC \cite{zhang2018deep}.

For the \textbf{Source and Target (ST)} setting, due to the lack of unsupervised domain adaptation methods for text-to-image person retrieval with available code, we select several state-of-the-art UDA baselines from related text-image multimodal tasks. These include POUF \cite{tanwisuth2023pouf} and ReCLIP \cite{hu2024reclip} for main experiments, and MAN \cite{jing2020cross}, ECN \cite{zhong2019invariance}, and ADDA \cite{tzeng2017adversarial} for intra-dataset experiments on CUHK-PEDES.

\subsection{Evaluation Metrics and Settings}\label{sec:metrics}

To quantitatively evaluate models in the cross-dataset text-to-image person retrieval task, we adopt two common metrics: Recall at Rank-$K$ (Rank-$K$) and Mean Average Precision (mAP), following existing works \cite{bai2023rasa, yan2023clip}. Rank-$K$ measures the proportion of target person images in the top $K$ ($K=1,5,10$) results, while mAP reflects the mean average precision of all results. Higher values for both metrics indicate better performance.

In the main cross-dataset experiments, we evaluate various baselines in the cross-dataset scenario for the above 3 datasets including ICFG-PEDES, RSTPReid, and CUHK-PEDES. 
For each dataset as the source set, we train the baselines on the source set and test them on the other two datasets respectively.
In the intra-dataset cross-domain experiments, we follow the cross-domain settings proposed by \cite{jing2020cross} within the CUHK-PEDES dataset. Specifically, we select SSM (S) as the source domain and consider 4 transfer tasks on CUHK03, Market-1501, VIPER, and CUHK01.

\begin{table*}[htbp]
    \centering
\begin{tabular}{cc|cc|cc|cc|cc}
\hline
\multirow{2}[0]{*}{Train Set} & \multirow{2}[0]{*}{Method} & \multicolumn{2}{c|}{S $\rightarrow$ C03} & \multicolumn{2}{c|}{S $\rightarrow$ M} & \multicolumn{2}{c|}{S $\rightarrow$ V} & \multicolumn{2}{c}{S $\rightarrow$ C01} \\
\cline{3-10}      &       & Rank-1 & Rank-5 & Rank-1 & Rank-5 & Rank-1 & Rank-5 & Rank-1 & Rank-5 \\
\hline
\multirow{6}[0]{*}{Source Only} & CMPM-CMPC  & 42.30  & 69.20  & 63.40  & 85.10  & 57.80  & 84.70  & 44.80  & 70.90  \\
      & MIA  & 49.00  & 76.70  & 66.20  & 86.20  & 55.10  & 84.70  & 50.20  & 75.90  \\
      & SCAN  & 50.20  & 75.90  & 64.20  & 86.20  & 55.10  & 81.10  & 48.20  & 76.80  \\
      & SSAN & 54.50  & 78.50  & 71.10  & 88.60  & 66.30  & 89.30  & 60.50  & 81.30  \\
      & EAIBC & 55.10  & 79.60  & 72.50  & 89.40  & 67.40  & 91.30  & 62.40  & 81.90  \\
      & RaSa & \underline{70.63}  & \underline{89.17}  & \underline{82.57}  & \underline{95.40}  & \underline{85.20}  & \underline{97.45}  & \underline{77.34}  & \underline{91.56}  \\
\hline
\multirow{5}[0]{*}{Source \& Target} & SPGAN & 44.70  & 72.50  & 63.30  & 85.30  & 60.70  & 85.70  & 45.30  & 71.20  \\
      & ADDA & 45.10  & 72.80  & 63.90  & 85.70  & 61.40  & 86.00  & 45.70  & 71.60  \\
      & ECN & 45.80  & 73.20  & 64.30  & 86.10  & 62.50  & 86.40  & 46.60  & 72.10  \\
      & MAN & 48.50  & 74.80  & 65.10  & 87.40  & 64.20  & 87.20  & 48.20  & 73.20  \\
      & \textbf{Ours}  & \textbf{71.25 } & \textbf{89.90 } & \textbf{83.54 } & \textbf{95.52 } & \textbf{86.73 } & \textbf{98.47 } & \textbf{78.91 } & \textbf{92.03 } \\
\hline
\end{tabular}%
    \caption{Comparison results with state-of-the-art methods on intra-dataset cross-domain settings within CUHK-PEDES dataset. The best results are emphasized in \textbf{bold}. The second-best results are noted by \underline{underline}.}
        \label{tab:extensive-study}%
\end{table*}%

\begin{table*}[htbp]
    \centering
\begin{tabular}{l|cccc|cccc}
    \hline
    \multicolumn{1}{c|}{\multirow{2}[0]{*}{Method}} & \multicolumn{4}{c|}{ICFG-PEDES $\rightarrow$ RSTPReid} & \multicolumn{4}{c}{ICFG-PEDES $\rightarrow$ CUHK-PEDES} \\
    \cline{2-9}      & Rank-1 & Rank-5 & Rank-10 & mAP   & Rank-1 & Rank-5 & Rank-10 & mAP \\
    \hline
    Baseline & 55.00  & 73.65  & 81.55  & 46.18   & 48.65  & 69.90  & 76.53  & 42.03 \\
    CMKD   & 58.20  & 78.85  & 85.25  & 48.50   & 52.04  & 72.42  & 79.46  & 45.33 \\
    \textbf{CMKD + GMP (Proposed)} & \textbf{59.95}  & \textbf{79.05}  & \textbf{85.95}  & \textbf{49.68}   & \textbf{52.70}  & \textbf{72.76}  & \textbf{80.15}  & \textbf{45.97} \\
    \hline
    \end{tabular}%
    \caption{Results of ablation study on ICFG-PEDES as the source dataset. The best results are emphasized in \textbf{bold}.}
    
        \label{tab:ablation-icfg}%
\end{table*}%

\subsection{Implementation Details}
The proposed model is implemented based on \textit{PyTorch 1.10} framework on Python 3.8 and Ubuntu 20.04. 
For each dataset, the image-text data is split according to the existing protocol \cite{bai2023rasa,yang2023towards}.
ALBEF \cite{li2021align} is adopted as the backbone of the vision language pre-training model and initialized with the pretraining weights on the source dataset.
During training, the batch size is set to $4$, and the optimizer is AdamW with an initial learning rate of $1e-5$ and cosine scheduler strategy. 
The hyperparameters of the proposed method are set as follows: the number of graph layers $L=2$, the number of neighbors $K=10$, the temperature $\tau=0.07$, the momentum coefficient $\alpha=0.999$, the loss coefficients $\lambda_1=\lambda_2=0.5, \lambda_3=1$.
All the experiments are conducted on NVIDIA GeForce RTX 4090.

\section{Result and Analysis}
In this section, we present the experimental results of our method and state-of-the-art baselines on the cross-dataset text-to-image person retrieval task. We also conduct ablation studies to analyze the effectiveness of each component.

\subsection{Comparison with State-of-the-arts}
To have a comprehensive evaluation of the proposed method, we compare it with state-of-the-art baselines on the cross-dataset text-to-image person retrieval task. As mentioned in Baselines, the compared baselines include the text-to-image person retrieval methods in the Source Only (SO) setting and the unsupervised domain adaptation methods in the Source and Target (ST) setting. The performance of the supervised training setting is also reported for reference.
The results on ICFG-PEDES as the source dataset are shown in Table \ref{tab:comparison-icfg}. The results on RSTPReid as the source dataset are shown in Table \ref{tab:comparison-rstp}. The results on the CUHK-PEDES in intra-dataset cross-domain settings are shown in Table \ref{tab:extensive-study}. 
From the results, we can make the following observations:

(1) The proposed method consistently outperforms the compared baselines on different transfer tasks. 
Compared with the second-best baseline, the proposed method achieves $4.37\%$ improvement of Rank-1 Recall and $3.29\%$ improvement of mAP on average.
The results demonstrate that the proposed method can effectively address the domain shift and modality heterogeneity challenges, and successfully transfer the knowledge learned from the source dataset to the target dataset for better text-to-image person retrieval performance.

(2) Existing single-domain text-to-image person retrieval methods generally suffer from significant performance degradation in cross-dataset scenarios. For example, the SOTA methods, APTM and RaSa, suffer an average performance drop of $25.30\%$ and $24.35\%$ in Rank-1 Recall respectively. This is mainly due to the challenge of data distribution differences between the source and target domains. The domain shift between different datasets makes it difficult for the single-domain model to transfer the knowledge and generalize well to the target domain.

(3) Existing unsupervised domain adaptation methods generally underperform. In the \textit{Source \& Target} setting, these baselines consistently lag behind single-domain retrieval baselines. The large image gallery scale exacerbates modality heterogeneity challenges in text-to-image person retrieval, limiting UDA method performance.

\subsection{Ablation Study}

To analyze the effectiveness of each component in the proposed method, we conduct ablation studies on ICFG-PEDES and CUHK-PEDES as the source dataset respectively. 
Specifically, we add each component to the baseline step by step, train the model according to the same settings as the main experiments, and evaluate the model on the target dataset.
Since the GMP module does not have any loss function and must rely on the training of the CMKD module, we evaluate the following three possible combinations:
\begin{itemize}
    \item \textbf{Baseline:} The backbone only.
    \item \textbf{CMKD:} The model with the CMKD module.
    \item \textbf{CMKD + GMP (Proposed):} The model with both the CMKD and GMP modules, i.e. the proposed GCKD.
\end{itemize}
The ablation study results are shown in Table \ref{tab:ablation-icfg}. 
As observed from the results, the baseline method achieves the lowest performance, which is due to the lack of domain adaptation. Adding the CMKD module to the baseline model improves the performance by a large margin, which demonstrates the effectiveness of the CMKD module in addressing the domain shift challenge. The proposed GMP module further improves the performance, which demonstrates the effectiveness of the proposed method in addressing the modality heterogeneity challenges. 
The results demonstrate the effectiveness of the proposed components in the cross-dataset text-to-image person retrieval task.

\section{Conclusion}
This paper presents a novel unsupervised domain adaptation method named Graph-Based Cross-Domain Knowledge Distillation (GCKD) for cross-dataset text-to-image person retrieval. In this method, a graph-based multi-modal propagation module and a contrastive knowledge distillation module are proposed to bridge the cross-domain correlation among the visual and textual samples and learn the cross-modal feature representation using the momentum knowledge distillation strategy. 
In the future, we plan to explore various advanced techniques to further improve cross-dataset retrieval performance, such as metric learning and adversarial learning.

\section{Acknowledgments}
This research is sponsored in part by the NSFC Program (No. U20A6003), Industrial Technology Infrastructure Public Service Platform Project "Public Service Platform for Urban Rail Transit Equipment Signal System Testing and Safety Evaluation" (No. 2022-233-225), Science and technology innovation project of Hunan Province (No.2023RC4014).

\bibliography{ref}

\end{document}